\documentclass[conference]{IEEEtran}
\IEEEoverridecommandlockouts
\usepackage{cite}
\usepackage{amsmath,amssymb,amsfonts}
\usepackage{algorithmic}
\usepackage{graphicx}
\usepackage{textcomp}
\usepackage{xcolor}
\usepackage{multirow}
\usepackage{subcaption}
\usepackage{mathtools}

\DeclarePairedDelimiter{\ceil}{\lceil}{\rceil}
\DeclarePairedDelimiter{\floor}{\lfloor}{\rfloor}
\def\BibTeX{{\rm B\kern-.05em{\sc i\kern-.025em b}\kern-.08em
    T\kern-.1667em\lower.7ex\hbox{E}\kern-.125emX}}
\begin{document}

\title{On-line Search History-assisted Restart Strategy for Covariance Matrix Adaptation Evolution Strategy
\thanks{This research was supported in part by a grant from the Hong Kong Research Grants Council under GRF Grant CityU 125313 and GRF Grant 11200317, the National Natural Science Foundation of China (Project No. 61603275, 61601329), and the Tianjin Higher Education Creative Team Funds Program.}
}

\author{\IEEEauthorblockN{Yang Lou, Shiu Yin Yuen, Guanrong Chen}
\IEEEauthorblockA{\textit{Department of Electronic Engineering} \\
\textit{City University of Hong Kong}\\
Hong Kong, China\\
felix.lou@my.cityu.edu.hk; \\ \{kelviny.ee; eegchen\}@cityu.edu.hk}

\and
\IEEEauthorblockN{Xin Zhang}
\IEEEauthorblockA{\textit{Tianjin Key Laboratory of Wireless} \\
\textit{Mobile Communications and Power Transmission} \\
%\textit{} \\	
\textit{Tianjin Normal University}\\
Tianjin, China\\
ecemark@tjnu.edu.cn}
}

\maketitle

\begin{abstract}
Restart strategy helps the covariance matrix adaptation evolution strategy (CMA-ES) to increase the probability of finding the global optimum in optimization, while a single run CMA-ES is easy to be trapped in local optima. In this paper, the continuous non-revisiting genetic algorithm (cNrGA) is used to help CMA-ES to achieve multiple restarts from different sub-regions of the search space. The CMA-ES with on-line search history-assisted restart strategy (HR-CMA-ES) is proposed. The entire on-line search history of cNrGA is stored in a binary space partitioning (BSP) tree, which is effective for performing local search. The frequently sampled sub-region is reflected by a deep position in the BSP tree. When leaf nodes are located deeper than a threshold, the corresponding sub-region is considered a region of interest (ROI). In HR-CMA-ES, cNrGA is responsible for global exploration and suggesting ROI for CMA-ES to perform an exploitation within or around the ROI. CMA-ES restarts independently in each suggested ROI. The non-revisiting mechanism of cNrGA avoids to suggest the same ROI for a second time. Experimental results on the CEC 2013 and 2017 benchmark suites show that HR-CMA-ES performs better than both CMA-ES and cNrGA. A positive synergy is observed by the memetic cooperation of the two algorithms.
\end{abstract}


\section{Introduction}

In the field of evolutionary computation, evaluation of solutions is considered as the most time consuming and computational resource consuming process. A well-designed evolutionary algorithm (EA) should cost negligible computational resource itself, compared to the cost of solution evaluations. Re-evaluation of any solution in EAs is wasteful and distorts the true performance of the EA\cite{Wolpert1997TEC}. 

In the real-valued problems, since the search space is defined in the continuous domain, the probability of generating two exactly the same solutions is rare. However, EAs do not search purely randomly in the space, and certain strategies in EAs may incur a risk of revisits. For example, it is observed that the \textit{absorbing scheme} of boundary handling method may generate revisiting solutions, especially when the global/local optima is on or near a boundary\cite{Lou2016MC}. The absorbing scheme sets the variable to the boundary value if it is generated outside the boundary. It is commonly adopted by artificial bee colony (ABC)\cite{Akay2012IS}, differential evolution (DE)\cite{Gandomi2012NCA}, and particle swarm optimization (PSO)\cite{Chu2011IS,Zambrano2013CEC}, etc, and thus, a non-negligible number of revisits on boundaries may exist in these EAs.

The non-revisiting genetic algorithm (NrGA)\cite{Yuen2009TEC} records the entire on-line search history by a binary space partitioning (BSP) tree, thus, re-evaluation of any solution is completely avoided. The continuous version cNrGA (where `c' stands for continuous) inherits the BSP tree to store the entire search history\cite{Chow2010CEC,Chow2012CEC}. Crossover is used in cNrGA for the first-stage searching, which introduces no new gene into the population, and thus identical solutions (revisits) are prone to be generated by recombinations. Selection further increases the probability of revisiting by giving the crossover opportunity to partially fit genes. Whenever a revisit happens, the chromosome (solution) would skip the re-evaluation, but instead, an adaptive mutation within a certain sub-region is performed. The adaptive mutation introduces new gene(s) into the population. More than just to avoid revisiting, the BSP tree archive naturally offers parameter-less adaptive mutation operations for local search.

The cNrGA with a proper memory management makes efficiently use of the entire on-line search history; cNrGA with least recently used (LRU) pruning (cNrGA-LRU) prunes the LRU leaves in the BSP tree, if the memory threshold is reached but the optimization is not terminated, and thus keeping the overall memory use constant. It is shown that cNrGA-LRU can maintain the performance of cNrGA, up to the empirical limit when 90\% of the search history is missing\cite{Lou2016MC}. 

The covariance matrix adaptation evolution strategy (CMA-ES)\cite{Hansen2006TNEC,Hansen2016arXiv} is a stochastic derivative-free optimizer which specializes in ill-conditioned local search. Though there is negligible or no revisit occurring in the search process of CMA-ES by nature\cite{Lou2016MC}, it is an optimizer that would frequently be trapped in local optima (premature convergence). An efficient remedy strategy for CMA-ES being trapped in local optima is to restart it\cite{Loshchilov2013CEC}. The niching strategy\cite{Preuss2010GECCO,Shir2010EC} for CMA-ES can be considered as a restart strategy that focuses on performing parallel (re)start, while a sequential restart strategy could consider when, where, and how. CMA-ES with restart strategies achieved great success in both CEC and BBOB competitions\cite{Loshchilov2013CEC}.

In this paper, a memetic algorithm of cNrGA and CMA-ES is proposed, namely CMA-ES with history-assisted restart strategy (HR-CMA-ES). HR-CMA-ES uses cNrGA as the global exploration method, and CMA-ES as the local exploitation method. The cooperation of the two algorithms bring benefit to each algorithm: It provides a powerful local optimizer for cNrGA, and it also provides CMA-ES good suggestions on region of interest (ROI) to restart, based on the search history.

On the other hand, instead of pruning a single memory unit as in cNrGA-LRU, HR-CMA-ES prunes a whole sub-region (i.e., ROI) if it has been exploited by CMA-ES. Thus, HR-CMA-ES will not suggest the same ROI for CMA-ES to exploit, but it does not forbid CMA-ES to exploit within the pruned sub-region again. The ROI suggested by cNrGA has two features: 1) the sub-region is actively visited by cNrGA; and 2) the sub-region has never been suggested before.

The rest of the paper is organized as follows: Sec. \ref{sec:pre} reviews cNrGA and CMA-ES. Sec. \ref{sec:hr} introduces the proposed HR-CMA-ES. Experimental study with comparisons and discussions, is given in Sec. \ref{sec:exp}. Conclusions are drawn in Sec. \ref{sec:con}.

\section{Preliminaries}
\label{sec:pre}

\subsection{cNrGA}
\label{cnrga}

\begin{figure}[htbp]
	\centering
	\includegraphics[width=.22\textwidth]{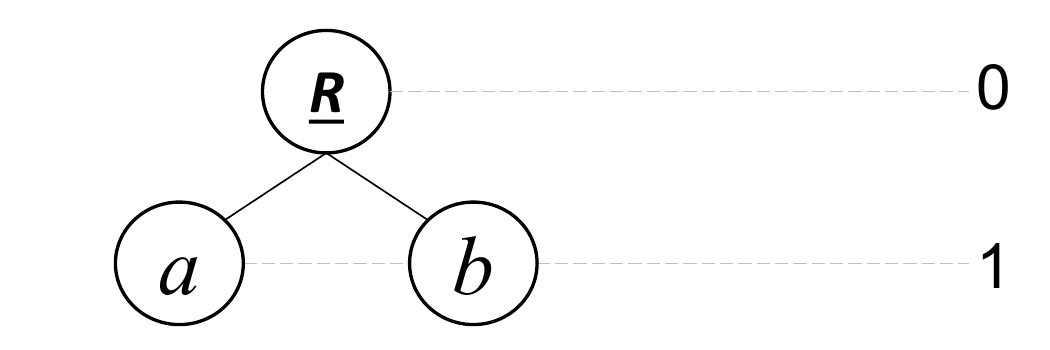}
	\caption{Two solutions $a$ and $b$ have been evaluated and inserted into the BSP tree. $R$ denotes the root node. Integers on the right denote the depth of tree nodes.}
	\label{fig2node}
\end{figure}

\begin{figure}[htbp]
	\centering
	\includegraphics[width=.22\textwidth]{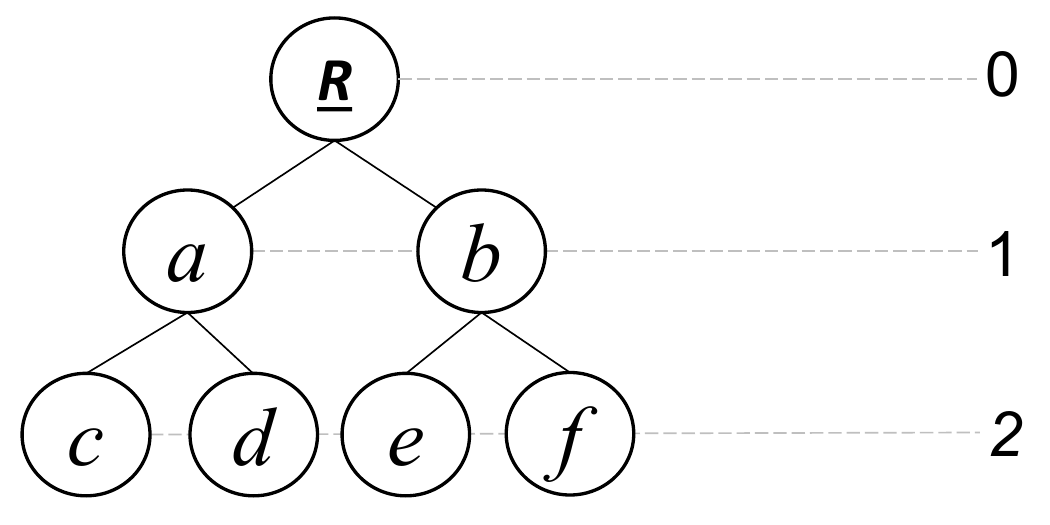}
	\caption{Four solutions have been evaluated and inserted in the BSP tree, where $a$ becomes a virtual node of $c$; and $b$ becomes a virtual node of $e$.}
	\label{fig4node}
\end{figure}

\begin{figure}[htbp]
	\centering
	\includegraphics[width=.48\textwidth]{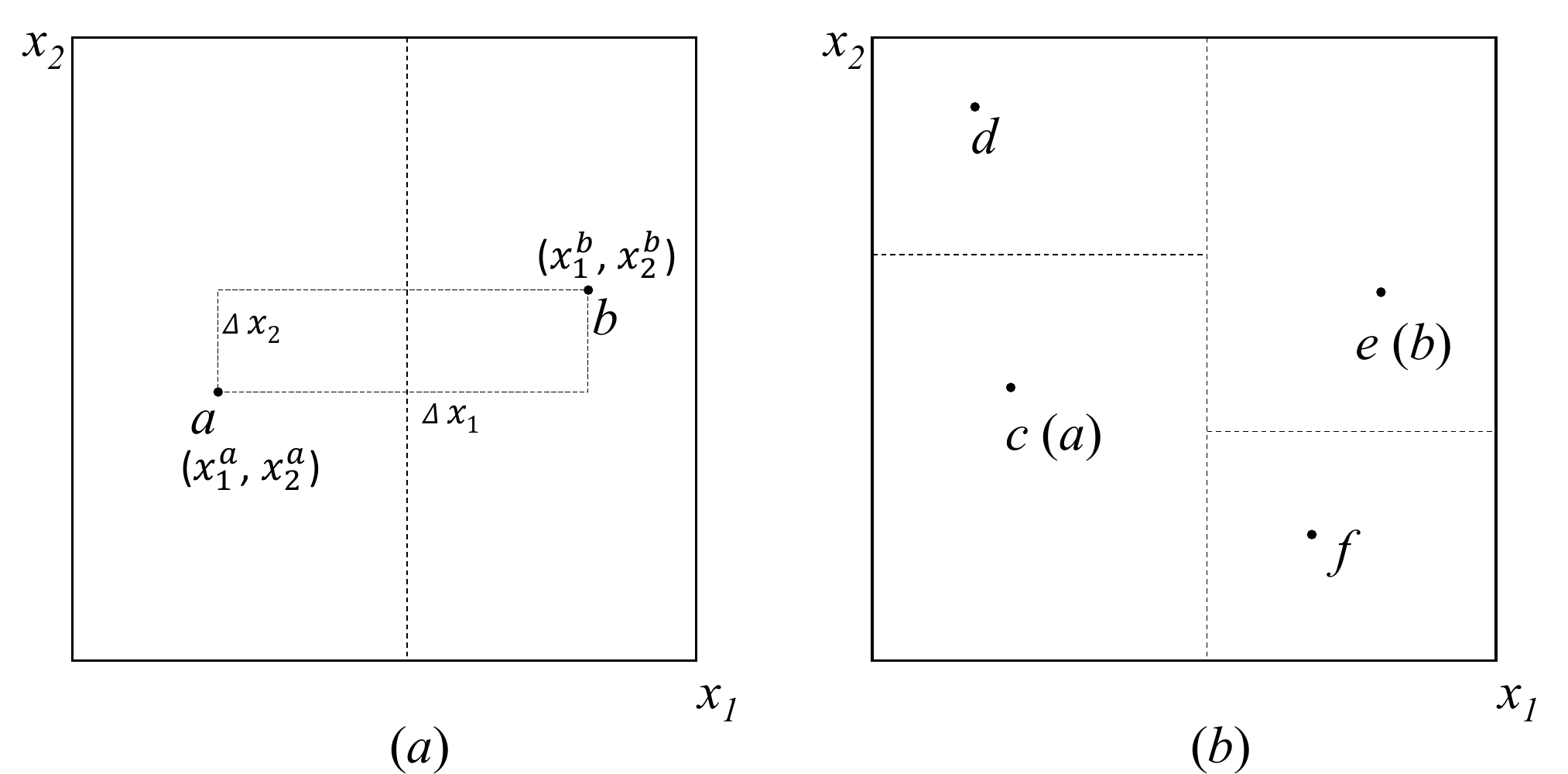}
	\caption{Search space partitioning: (a) two solutions $a$ and $b$ partition the search space into 2 sub-regions; (b) four solutions partition the space into 4 parts.}
	\label{figspace1}
\end{figure}

Search history is accumulatively stored at each generation in cNrGA: The BSP tree expands and the whole search space $S$ is partitioned. Each evaluated solution is represented by a leaf node in the BSP tree, which occupies a sub-region in the partitioned space. Suppose the current number of evaluated solutions is $M$, then $S$ is divided into $M$ sub-regions. For example, two solutions partition the search space into two parts, and the partitioning line is drawn on the dimension along which the two solutions have the maximum difference. Figs. \ref{fig2node}, \ref{fig4node} and \ref{figspace1} show an example of BSP tree expansion and search space partitioning in a 2-dimensional space.

Figure \ref{fig2node} shows when there are two evaluated solutions $a$ and $b$, the BSP tree has two leaf nodes at depth 1 (depth 0 represents the root node). Accordingly, in Fig. \ref{figspace1} (a) the 2-dimensional space is divided into two regions. Since the two solutions $a$ and $b$ have a greater difference along the dimension $x_1$ than $x_2$ ($\Delta x_1=|x_1^a-x_1^b| > \Delta x_2=|x_2^a-x_2^b|$), the partitioning line is drawn at the center of $x_1^a$ and $x_1^b$, on the dimension $x_1$. 

As the number of evaluated solutions increases, more nodes are inserted into the BSP tree as shown in Fig. \ref{fig4node}; and the search space $S$ is further divided into smaller regions as shown in Fig. \ref{figspace1} (b). There are now four evaluated solutions as shown in Fig. \ref{fig4node}, which are all leaf nodes located in depth 2. Nodes $a$ and $c$ store the same evaluated solution, but node $c$ has a smaller sub-region size than node $a$ in the search space (i.e., nodes $c$ and $d$ divide the sub-region of $a$). Node $a$ becomes a virtual node of node $c$, which is an intermediate node to reach other leaf nodes.

\subsection{CMA-ES}
\label{cmaes}

At each iteration of CMA-ES, new solutions are sampled according to a multivariate normal distribution, as shown in Eq. \ref{eqcma}, where $x_{i,g+1}$ denotes the $i$th solution at generation $g+1$. Solutions with good fitness values survive and are used to update the center of the multivariate normal distribution ($m_g$) for the next generation sampling. The movement of center (towards region with good fitness) is used to update the evolution paths, including the isotropic and anisotropic path. The isotropic evolution path is then used to update the step size ($\sigma_g$); the anisotropic path is used to update the covariance matrix ($C_g$).

\begin{equation}
\label{eqcma}
	x_{i,g+1}\sim\mathcal{N}(m_g,\sigma_{g}^2C_{g})\sim m_g+\sigma_{g}\times \mathcal{N}(0,C_{g})
\end{equation}

CMA-ES is particularly powerful if the problem is ill-conditioned, but is weak if the problem landscape is rugged, when the covariance matrix may be trapped in a local optima. As the searching process converges, the region near the center of the multivariate normal distribution ($m_g$) has a much higher probability of being sampled, while the region far away from the center may not be sufficiently explored. Meanwhile, CMA-ES itself does not have a mechanism to escape from local optima. However, the probability of CMA-ES being stuck in local optima can be lowered when it is restarted, and thus the restart strategy helps CMA-ES to become a good global optimizer, rather than a local optimizer. 

CMA-ES has developed a set of stopping criteria: Before the maximum number of evaluations is used up, CMA-ES would stop, for example, if the condition number of covariance matrix is large (e.g., greater than $1\times 10^{14}$), or if the best objective values is not changed in the last several, say $10+\ceil{30D/\lambda}$ ($D$ and $\lambda$ represent problem dimension and population size, respectively), generations. In addition, successively increased population sizes in CMA-ES would benefit exploration\cite{Loshchilov2013CEC}. The stopping criteria could answer the \textit{when to restart} question; the population size modification suggests \textit{how to restart}, but neither considers \textit{where to restart}.

\section{Search History-Based Restart Strategy}
\label{sec:hr}

The proposed HR-CMA-ES is an attempt to answer the question of \textit{where to restart} CMA-ES. A preferable strategy is to restart from an unexplored position, which requires a full memory of search history. The non-revisiting scheme records the entire search history to avoid revisiting in the discrete domain, and to perform adaptive mutation in both the discrete and continuous domains. Therefore, the non-revisiting scheme is suitable to be employed to suggest the region to (re)start CMA-ES.

As illustrated in Sec. \ref{cnrga}, a BSP tree of depth $\tau$ is capable of recording $2^{\tau}$ solutions, only if the BSP tree is balanced. Practically, the formed BSP tree is not balanced. Suppose there are $2^{\tau}$ solutions in the BSP tree; many leaf nodes are located deeper than $\tau$, while many nodes are shallower than $\tau$. The solutions that are deeper than $\tau$ suggest a region of interest (ROI) for exploitation, while the shallow ones suggest relatively less promising sub-regions. For example, the least recently used solutions in cNrGA-LRU\cite{Lou2016MC} are found amongst the shallow solutions in the BSP tree.

\begin{figure}[htbp]
	\centering
	\includegraphics[width=.26\textwidth]{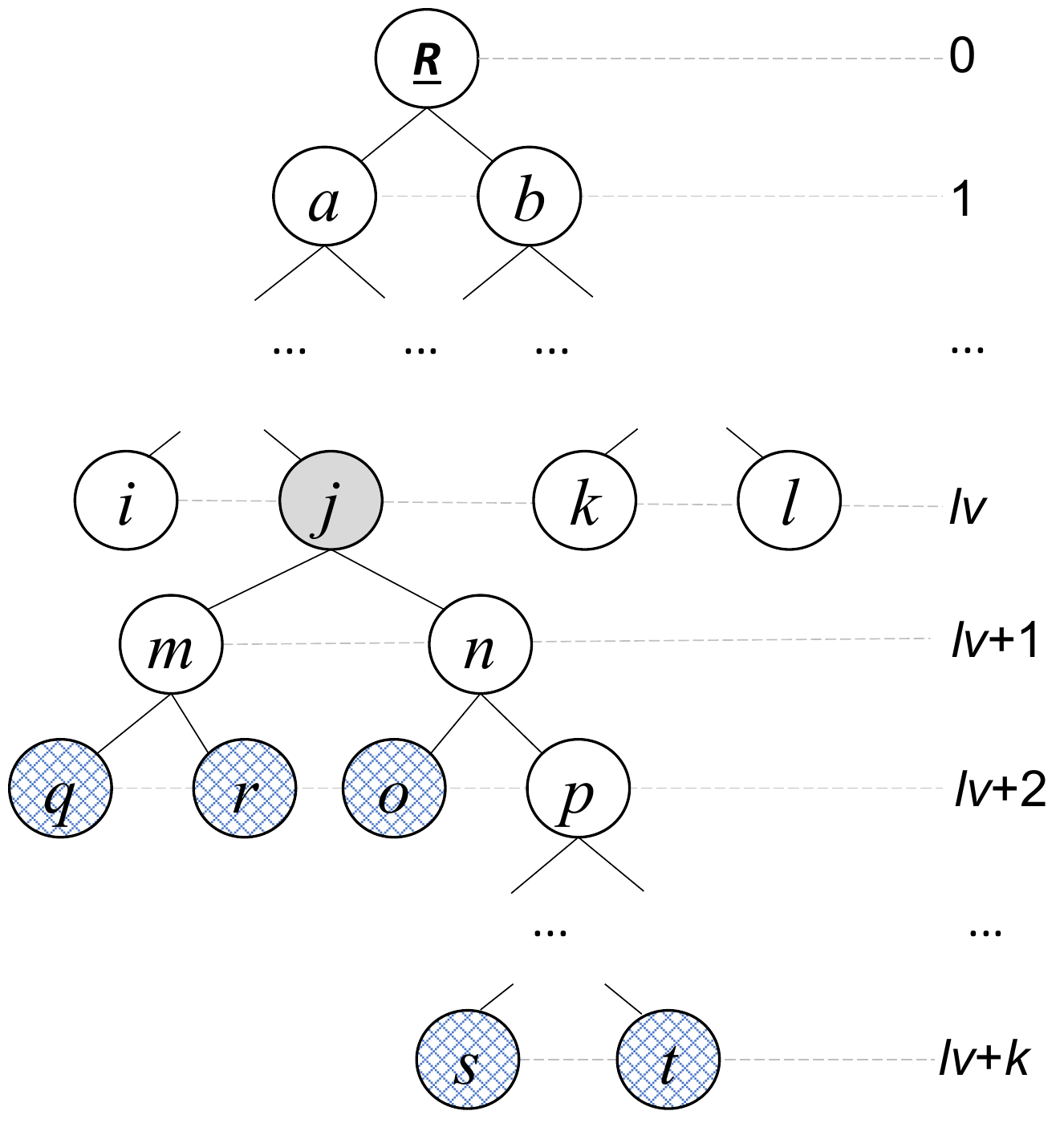}
	\caption{Once any leaf node reaches a depth of $lv+k$, then CMA-ES is triggered: The traced-back node on depth $lv$ (i.e., node $j$) is regarded as the root of the sub-tree, and all the leaf nodes under $j$ form the initial population of CMA-ES.}
	\label{figlv+k}
\end{figure}

Suppose the solutions are allocated in a balanced manner in the BSP tree. Given the maximum number of fitness evaluations is $M_e$, the depth of a BSP tree to record the entire search history is $lv=\ceil{\log_{2}{M_e}}$, where $\ceil{x}$ represents the minimum integer that is greater than or equal to the real number $x$. Thus, $lv$ is regarded as the average depth of the BSP tree. Any solution that exploits deeper than $lv$ may suggest a more promising sub-region than average for further exploitation. Fig. \ref{figlv+k} shows an example, where nodes $s$ and $t$ have reached the depth $lv+k$. Then, cNrGA would be suspended and CMA-ES would be triggered next. Node $j$ is the root of the sub-tree that contains nodes $s$ and $t$. All the leaf nodes under node $j$ are considered solutions within an ROI, and employed as the initial population for CMA-ES restart. The recommended population size ($\lambda$) of CMA-ES is shown in Eq. \ref{eqlambda}, where $D$ represents the problem dimension. 

\begin{equation}
	\label{eqlambda}
	\lambda=4+ \floor{3\times \ln{D}}
\end{equation}

In HR-CMA-ES, to collect approximately $\lambda$ solutions for the CMA-ES initialization, a sub-tree with depth $k=\ceil{\log_{2}{\lambda}}$ under the sub-root (at depth $lv$) is needed. Note that a sub-tree with depth $k$ may include $\lambda_0$ ($k+1\leq\lambda_0\leq\lambda$) solutions. If any leaf node reaches the depth of $lv+k$, all the leaf nodes in the sub-tree (with the sub-root at depth $lv$) are collected as the initial population of CMA-ES, and meanwhile the running of cNrGA is suspended. As shown in Fig. \ref{figlv+k}, node $j$ at depth $lv$ is the sub-root; all the leaf nodes under node $j$, namely, nodes $q$, $r$, $o$, $s$, $t$, etc. are employed as the initial population of CMA-ES restart.

\begin{figure}[htbp]
	\centering
	\includegraphics[width=.26\textwidth]{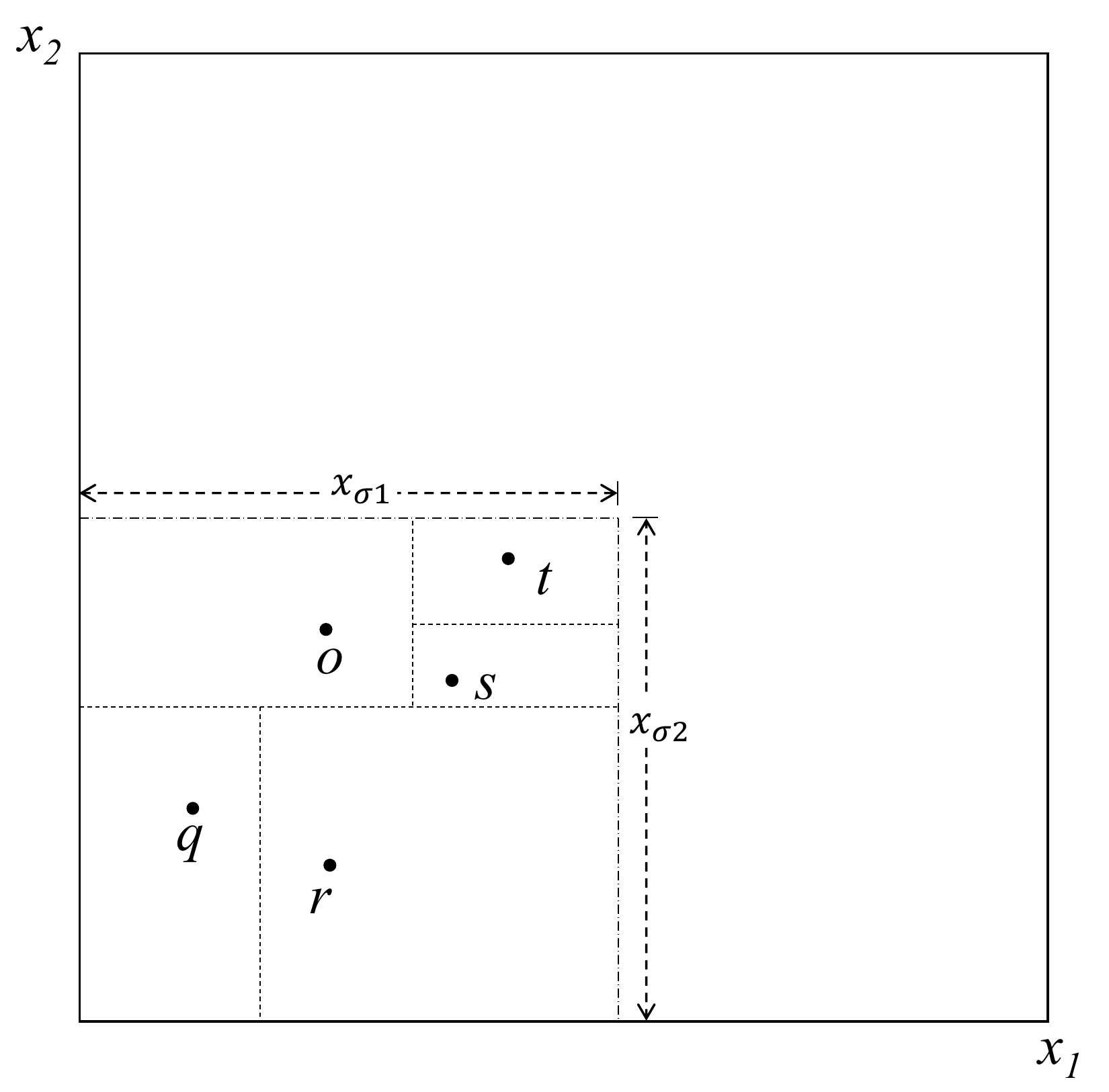}
	\caption{The corresponding sub-region in the search space. The sub-region of size: $x_{\sigma1}\times x_{\sigma2}$ is the sub-region for node $j$ (the entire sub-region of the sub-trees); as it is further partitioned, each divided piece represents a leaf node.}
	\label{figlv_subregion}
\end{figure}

\begin{figure}[htbp]
	\centering
	\includegraphics[width=.26\textwidth]{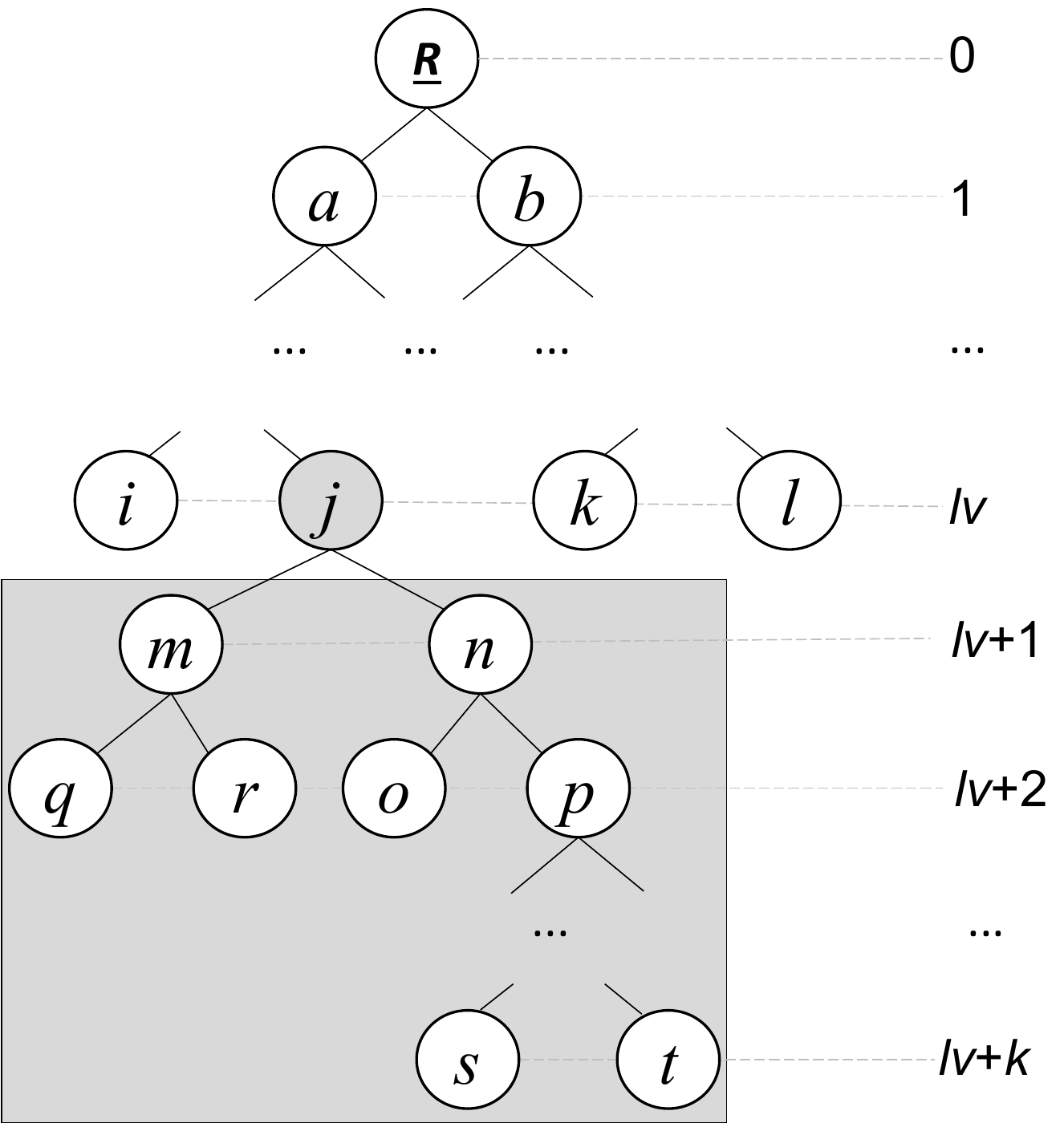}
	\caption{After CMA-ES exploitation, the entire sub-region is regarded a blocked region for cNrGA, thus, any solution generated by cNrGA within the sub-region would be considered a revisit, and discarded without evaluation. Instead, another random solution would be generated outside the blocked region.}
	\label{figbsp_blk1}
\end{figure}

\begin{figure}[htbp]
	\centering
	\includegraphics[width=.26\textwidth]{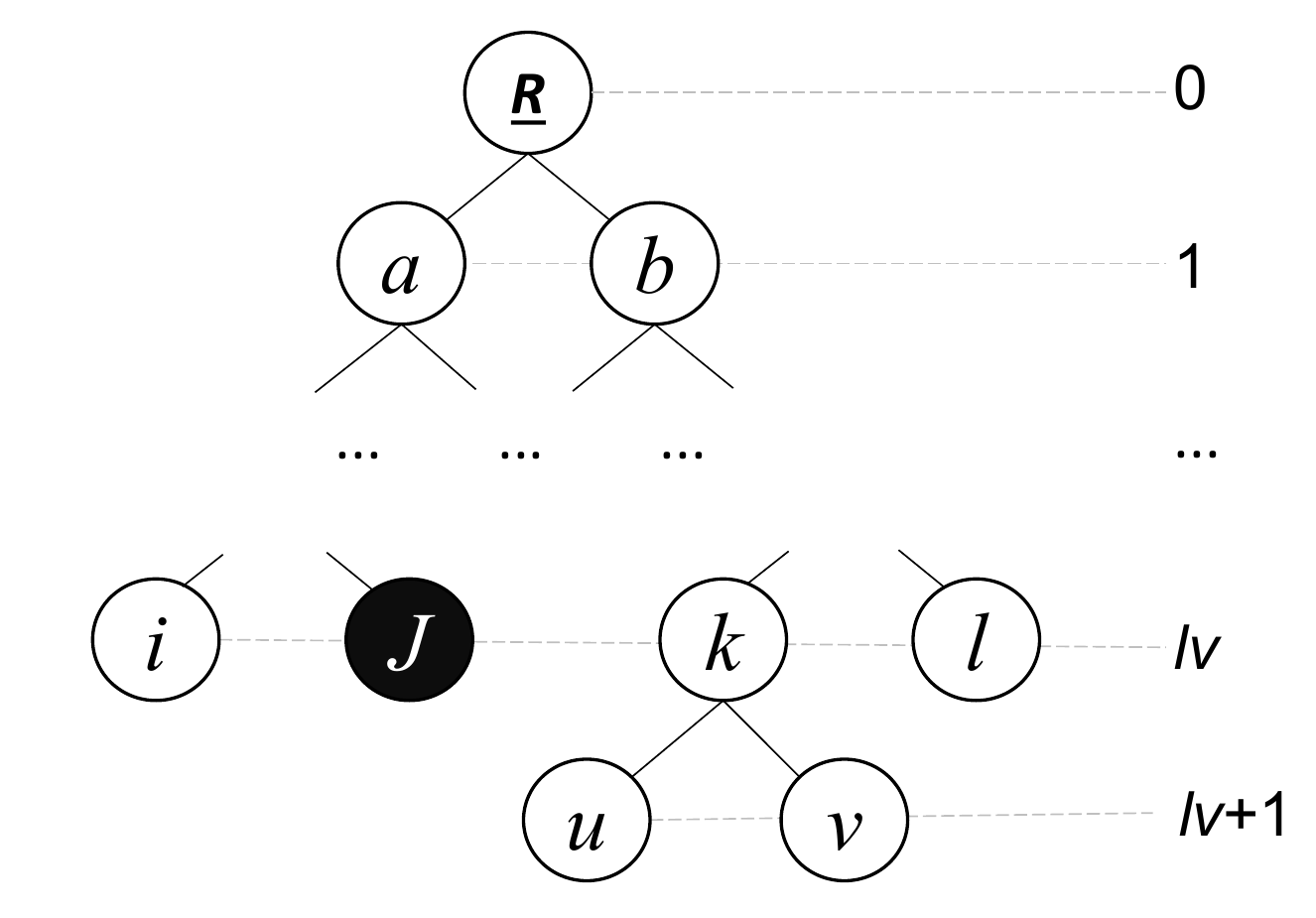}
	\caption{When the tree node is blocked, no further insertion is allowed under the node. In contrast, other regions would be further exploited, if any of them becomes an ROI.}
	\label{figbsp_blk2}
\end{figure}

\begin{figure}[htbp]
	\centering
	\includegraphics[width=.48\textwidth]{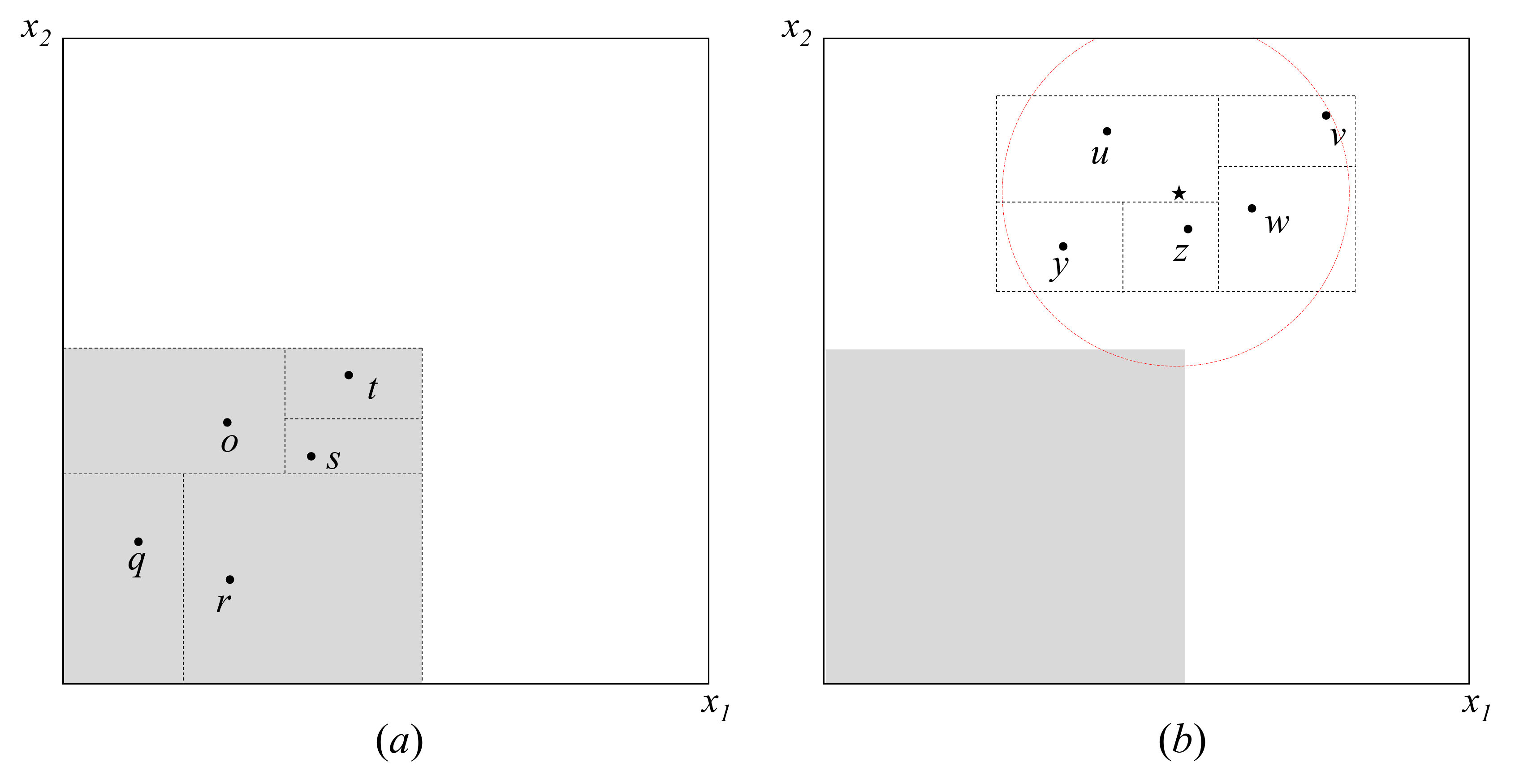}
	\caption{(a) The gray-shaded sub-region has been blocked to cNrGA. (b) The cNrGA explores a new ROI, and then CMA-ES is restarted where. The gray-shaded sub-region is not closed to CMA-ES.}
	\label{figbsp_blk_subregion}
\end{figure}

The mean of all these collected solutions is used as the initial center position of CMA-ES. The initial step size is set to $\sigma=0.3\times\max\{x_{\sigma1},x_{\sigma2}\}$. The CMA-ES takes over the optimization until any one of its stopping criteria is met. After CMA-ES is run, the sub-region is denoted a blocked region for cNrGA (see Figs. \ref{figbsp_blk1} and \ref{figbsp_blk2}). Note that this sub-region is not blocked to CMA-ES, though CMA-ES may have very low probability visiting it again. As illustrated in Fig. \ref{figbsp_blk_subregion}), CMA-ES is allowed to sample within the sub-region that is blocked to cNrGA.

\begin{figure}[htbp]
	\centering
	\includegraphics[width=.34\textwidth]{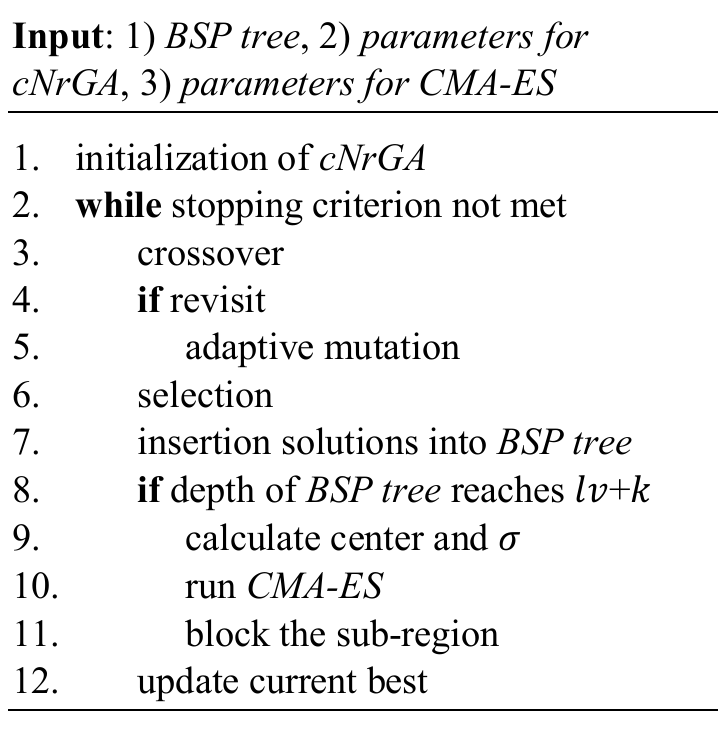}
	\caption{Pseudo code for HR-CMA-ES.}
	\label{pseudocode}
\end{figure}

\section{Experimental Study}
\label{sec:exp}

\subsection{Benchmark Comparison}

The CEC 2013\cite{CEC2013} and CEC 2017\cite{CEC2017} bound constrained real-valued benchmark suites are employed for testing. There are 28 problems in CEC 2013 suite (denoted by f1301 to f1328), and 29 problems in CEC 2017 suite (denoted by f1701, f1703 to f1730, where f1702 is excluded because it shows unstable behavior\cite{CEC2017}). The problem dimension is set to 10D and 30D. The maximum number of problem evaluations is $1\times 10^5$ for 10D problems and $3\times 10^5$ for 30D problems. The number of independent runs is set to 30. The significance test is performed by the Kruskal-Wallis test\cite{Lou2018GECCO} with confidence level $\alpha=0.05$.

Two algorithms, CMA-ES and cNrGA-LRU, are compared with HR-CMA-ES. All three algorithms are implemented in Matlab. The source code of CMA-ES is downloaded from \cite{CMA_Matlab} and cNrGA from\cite{NR_Matlab}. The employed CMA-ES is a version 3.33 copy with default parameter settings. Before the maximum number of evaluations is used up, CMA-ES may stop and restart, if any of the termination criteria is met. The recommended termination criteria are presented in Sec. B.3 of \cite{Hansen2016arXiv}. The population size of cNrGA-LRU is 100 and the crossover rate is 0.5 as recommended in\cite{Lou2016MC}. The memory-pruning rate of cNrGA-LRU is 0.5, meaning that half of the search history that is recognized as LRU information is discarded during the optimization.

% --- Table #1 CEC 2013 --- %
\begin{table}[htbp]
	\begin{center}
		\caption{Comparison of the three EAs on CEC 2013 benchmarks. (+) represents the algorithm is superior to HR-CMA-ES, and (--) represents inferior to HR-CMA-ES.}
		\label{tb13}
		\begin{tabular}{c|ccc|ccc}
			\hline
			\multirow{2}{*}{\begin{tabular}[c]{@{}c@{}}CEC\\ 2013\end{tabular}} & \multicolumn{3}{c|}{10D} & \multicolumn{3}{c}{30D}	\\ \cline{2-7}
		   		 & HR    & CMA   & LRU   & HR    & CMA   & LRU    \\ \hline
			f1301& 1.5   & 1.5   & 3 (--) & 1.5   & 1.5   & 3 (--)  \\
			f1302& 1.5   & 1.5   & 3 (--) & 1.5   & 1.5   & 3 (--)  \\
			f1303& 1.5   & 1.5   & 3 (--) & 2     & 1     & 3 (--)  \\
			f1304& 1.5   & 1.5   & 3 (--) & 1.5   & 1.5   & 3 (--)  \\
			f1305& 1.5   & 1.5   & 3 (--) & 1.5   & 1.5   & 3 (--)  \\ \hline
			f1306& 1.5   & 1.5   & 3 (--) & 1.5   & 1.5   & 3 (--)  \\
			f1307& 2     & 1     & 3 (--) & 1     & 2     & 3 (--)  \\
			f1308& 2     & 3 (--)& 1 (+)  & 2     & 3 (--)& 1 (+)  \\
			f1309& 1     & 2     & 3 (--) & 2     & 1     & 3 (--)  \\
			f1310& 1.5   & 1.5   & 3 (--) & 1     & 2     & 3 (--)  \\
			f1311& 2     & 3 (--) & 1 (+) & 2     & 3 (--) & 1 (+)  \\
			f1312& 2     & 1     & 3 (--) & 2     & 1     & 3 (--)  \\
			f1313& 2     & 1     & 3 (--) & 2     & 1     & 3 (--)  \\
			f1314& 2     & 3 (--) & 1 (+) & 2     & 3 (--) & 1 (+)  \\
			f1315& 1     & 2     & 3 (--) & 1     & 2     & 3 (--)  \\
			f1316& 2     & 3 (--) & 1     & 2     & 3 (--) & 1 (+)  \\
			f1317& 2     & 3     & 1 (+) & 2     & 3     & 1 (+)  \\
			f1318& 1     & 2     & 3 (--) & 1     & 2     & 3 (--)  \\
			f1319& 2     & 3     & 1     & 2     & 3     & 1 (+)  \\
			f1320& 2     & 3     & 1 (+) & 2     & 3 (--) & 1 (+)  \\ \hline
			f1321& 2     & 1     & 3 (--) & 2     & 1     & 3 (--)  \\
			f1322& 2     & 3 (--) & 1 (+) & 2     & 3 (--) & 1 (+)  \\
			f1323& 1     & 2     & 3     & 2     & 1     & 3 (--)  \\
			f1324& 2     & 1 (+) & 3 (--) & 2     & 1 (+) & 3 (--)  \\
			f1325& 2     & 1     & 3 (--) & 2     & 1     & 3 (--)  \\
			f1326& 2     & 1     & 3 (--) & 3     & 1 (+) & 2      \\
			f1327& 2     & 1 (+) & 3 (--) & 2     & 1 (+) & 3 (--)  \\
			f1328& 2     & 1     & 3     & 2     & 1 (+) & 3 (--)  \\ \hline
			avg  & 1.73  & 1.84  & 2.43  & 1.80  & 1.80  & 2.39   \\ \hline
			(+)  &       & 2     & 6     &       & 4     & 8      \\
			(--) &       & 4     & 18    &       & 6     & 19     \\ \hline
		\end{tabular}
	\end{center}
\end{table}
% --- Table #1 CEC 2013 --- %

Table \ref{tb13} shows the rank comparison of the three algorithms on solving the CEC 2013 problems. If two or more algorithms obtain the same result, then they share the rank. For example, HR-CMA-ES and CMA-ES get the same result for f1301, and thus they share the first and second ranks, with each algorithm obtaining rank 1.5. Note that although the significance test is multi-sample wise, the significance mark (+/--) is marked pair-wise for the algorithm compared to HR-CMA-ES. For the 10D problems, both HR-CMA-ES and CMA-ES outperform cNrGA-LRU significantly on solving the first 5 unimodal problems (f1301 to f1305). For the 15 basic multimodal problems (f1306 to f1320), it is observed that cNrGA-LRU performs bipolarly. Specifically, cNrGA-LRU outperforms HR-CMA-ES significantly on solving 5 problems (f1308, f1311, f1314, f1317, and f1320), while it is significantly worse than HR-CMA-ES on solving 8 problems (f1306, f1307, f1309, f1310, f1312, f1313, f1315, and f1318). CMA-ES performs significantly worse than HR-CMA-ES on solving 4 problems (f1308, f1311, f1314, and f1316), and without outperforming HR-CMA-ES on any basic multimodal problem.

As for the last 8 composition problems (f1321 to f1328), CMA-ES is superior to HR-CMA-ES on solving problems f1324 and f1327, but is inferior on f1322. In contrast, cNrGA-LRU performs well on f1322, but badly on problems f1324 to f1327. The overall performance of HR-CMA-ES on 10D problems is that 1) the average rank obtained by HR-CMA-ES is the best over the three algorithms; and 2) HR-CMA-ES is inferior to CMA-ES and cNrGA-LRU on solving 2 and 6 problems respectively, but is superior to them on 4 and 18 problems, respectively. Similar ranks and significance test results are observed on solving the 30D problems.

The detailed results of HR-CMA-ES and CMA-ES on solving CEC 2013 problems are given in Table \ref{tbres_cec13_10d} (for 10D problems) and Table \ref{tbres_cec13_30d} (for 30D problems).

% --- Table #2 CEC 2017 --- %
\begin{table}[htbp]
	\begin{center}
		\caption{Comparison of the three EAs on CEC 2017 benchmarks. (+) represents the algorithm is superior to HR-CMA-ES, and (--) represents inferior to HR-CMA-ES.}
		\label{tb17}
		\begin{tabular}{c|ccc|ccc}
			\hline
			\multirow{2}{*}{\begin{tabular}[c]{@{}c@{}}CEC\\ 2017\end{tabular}} & \multicolumn{3}{c|}{10D} & \multicolumn{3}{c}{30D} \\ \cline{2-7}
		   		 & HR   & CMA    & LRU   & HR   & CMA    & LRU    \\ \hline
			f1701& 1.5   & 1.5   & 3 (--) & 1.5   & 1.5   & 3 (--)  \\
			f1703& 1.5   & 1.5   & 3 (--) & 1.5   & 1.5   & 3 (--)  \\ \hline
			f1704& 1.5   & 1.5   & 3 (--) & 2     & 1     & 3 (--)  \\
			f1705& 2     & 1     & 3 (--) & 2     & 1     & 3      \\
			f1706& 1     & 2     & 3 (--) & 2     & 3     & 1      \\
			f1707& 2     & 1     & 3 (--) & 2     & 1     & 3 (--)  \\
			f1708& 1     & 2     & 3 (--) & 1     & 2     & 3 (--)  \\
			f1709& 1.5   & 1.5   & 3 (--) & 2     & 1     & 3 (--)  \\
			f1710& 2     & 3     & 1     & 1     & 3     & 2      \\ \hline
			f1711& 2     & 3 (--) & 1     & 1     & 2 (--) & 3 (--)  \\
			f1712& 1     & 2     & 3 (--) & 2     & 1     & 3 (--)  \\
			f1713& 1     & 2     & 3 (--) & 2     & 1     & 3 (--)  \\
			f1714& 1     & 2     & 3 (--) & 1     & 2     & 3 (--)  \\
			f1715& 1     & 2     & 3 (--) & 2     & 1     & 3 (--)  \\
			f1716& 1     & 2     & 3 (--) & 1     & 2     & 3 (--)  \\
			f1717& 1     & 2     & 3     & 1     & 2 (--) & 3 (--)  \\
			f1718& 1     & 2     & 3 (--) & 1     & 2     & 3 (--)  \\
			f1719& 2     & 1     & 3 (--) & 1     & 2     & 3 (--)  \\
			f1720& 2     & 3 (--) & 1 (+) & 1     & 2     & 3      \\ \hline
			f1721& 3     & 1     & 2     & 2     & 1     & 3 (--)  \\
			f1722& 2     & 1     & 3 (--) & 1     & 3 (--) & 2 (--)  \\
			f1723& 2     & 1 (+) & 3 (--) & 2     & 1 (+) & 3      \\
			f1724& 2     & 1 (+) & 3 (--) & 3     & 1 (+) & 2      \\
			f1725& 2     & 1     & 3 (--) & 1     & 2     & 3 (--)  \\
			f1726& 2     & 1     & 3 (--) & 2     & 1 (+) & 3 (--)  \\
			f1727& 1     & 2 (--) & 3 (--) & 1     & 2     & 3 (--)  \\
			f1728& 2     & 1     & 3 (--) & 1     & 2     & 3 (--)  \\
			f1729& 1     & 3     & 2     & 1     & 3     & 2      \\
			f1730& 2     & 1     & 3 (--) & 1     & 2     & 3 (--)  \\ \hline
			avg& 1.59  & 1.69  & 2.72  & 1.48  & 1.72  & 2.79   \\ \hline
			(+)&       & 2     & 1     &       & 3     & 0      \\
			(--)&       & 3     & 23    &       & 3     & 22     \\ \hline
		\end{tabular}
	\end{center}
\end{table}
% --- Table #2 CEC 2017 --- %

Table \ref{tb17} shows the rank comparison of the three algorithms on solving the CEC 2017 benchmark suite, 10D and 30D. As in the CEC 2013 comparison, both HR-CMA-ES and CMA-ES outperform cNrGA-LRU on solving the 2 unimodal problems (f1701 and f1703). For the following 7 multimodal problems (f1704 to f1710), a similar phenomenon as in CEC 2013 is observed, i.e., cNrGA-LRU either performs the best (f1710-10D and f1706-30D), or performs the worst (the rest of multimodal problems except f1710-30D). In contrast, HR-CMA-ES and CMA-ES performs close to each other on these problems.

Problems f1711 to f1720 are hybrid problems, which are difficult for cNrGA-LRU to handle. HR-CMA-ES performs quite well on solving these problems, not only because it gains many rank one results, but also because it performs significantly better than CMA-ES in 4 cases (f1711-\{10D,30D\}, f1717-30D, and f1720-10D). 

The last 10 problems in CEC 2017 suite (f1721 to f1730) are composition problems. Though the ranks of HR-CMA-ES and CMA-ES are close to each other, CMA-ES gains more significantly better results on solving these problems. CMA-ES performs better on the 10D problems, while HR-CMA-ES performs better on the 30D problems. 

In summary, HR-CMA-ES gains the best overall ranks on the CEC 2017 test. Compared to CMA-ES, HR-CMA-ES gains one more superior results to CMA-ES than inferior to CMA-ES. Compared to cNrGA-LRU, HR-CMA-ES is inferior to cNrGA-LRU only once, but superior to cNrGA-LRU on 23 10D problems and on 22 30D problems, respectively.

The detailed results of HR-CMA-ES and CMA-ES on solving CEC 2017 problems are given in Table \ref{tbres_cec17_10d} (for 10D problems) and Table \ref{tbres_cec17_30d} (for 30D problems).

\subsection{Discussions}

History-assisted restart strategy improves the performance of CMA-ES on solving multimodal problems. For the 32 multimodal and hybrid problems (i.e., f1306 to f1320, f1704 to f1720), HR-CMA-ES performs significantly better than CMA-ES on 6 10D problems (f1308, f1311, f1314, f1316, f1711, and f1720) and 7 30D problems (f1308, f1311, f1314, f1316, f1320, f1711, and f1717), without performing significantly worse than CMA-ES.

In this paper, HR-CMA-ES is considered as a CMA-ES with search history-assisted restart, i.e., cNrGA explores a suitable promising ROI for CMA-ES to (re)start. Once CMA-ES has finished an exploitation in the ROI, it shuts down the opportunity of cNrGA suggesting it again. On the other hand, HR-CMA-ES could also be considered a novel cNrGA with CMA-ES as a powerful local search. As soon as an ROI is explored by cNrGA, a more powerful local optimizer CMA-ES is applied, instead of the adaptive mutation. Once exploited, the sub-region is blocked to cNrGA, which is equivalent to a novel pruning strategy that prunes a whole sub-region, rather than a single solution. It is a more efficient pruning strategy in the continuous space. A positive algorithm synergy\cite{Yuen2016ASC} is observed in the cooperation of cNrGA and CMA-ES in this work.

Compared to other CMA-ES variants with restart, HR-CMA-ES suggests \textit{where} to (re)start, but is unable to suggest \textit{how} to (re)start. For example, strategy like population size modification is difficult to be applied in HR-CMA-ES, since CMA-ES is (re)initiated in different sub-regions of the search space, but the suitable population sizes for different sub-regions are difficult to predict.

\section{Conclusions}
\label{sec:con}

The covariance matrix adaptation evolution strategy (CMA-ES) is prone to be trapped in local optima if the problem landscape being solved is rugged. Restart strategy turns out to be an efficient remedy to help CMA-ES escape from being trapped. The continuous non-revisiting genetic algorithm (cNrGA) utilizes the entire on-line search history to perform revisiting avoidance and adaptive mutation. Since the question about \textit{when} and \textit{how} to restart CMA-ES has been well studied, this paper attempts to answer \textit{where} to restart. The history-based restart CMA-ES (HR-CMA-ES) is proposed. cNrGA suggests region of interest (ROI) for CMA-ES to perform exploitation. An ROI is a sub-region where cNrGA generates solutions frequently. The non-revisiting mechanism avoids suggesting the same ROI for CMA-ES repeatedly. Experimental results on the CEC 2013 and 2017 benchmark suites show that HR-CMA-ES significantly outperforms cNrGA with constant memory, and performs better than CMA-ES in terms of average rank. Particularly, HR-CMA-ES shows powerful optimization ability on solving multimodal problems. A positive synergy is observed by the cooperation of the two algorithms, which makes HR-CMA-ES a better algorithm than the two component algorithms.

\bibliographystyle{IEEEtran}
\bibliography{IEEEabrv,myref}

\begin{thebibliography}{10}
\providecommand{\url}[1]{#1}
\csname url@samestyle\endcsname
\providecommand{\newblock}{\relax}
\providecommand{\bibinfo}[2]{#2}
\providecommand{\BIBentrySTDinterwordspacing}{\spaceskip=0pt\relax}
\providecommand{\BIBentryALTinterwordstretchfactor}{4}
\providecommand{\BIBentryALTinterwordspacing}{\spaceskip=\fontdimen2\font plus
\BIBentryALTinterwordstretchfactor\fontdimen3\font minus
  \fontdimen4\font\relax}
\providecommand{\BIBforeignlanguage}[2]{{%
\expandafter\ifx\csname l@#1\endcsname\relax
\typeout{** WARNING: IEEEtran.bst: No hyphenation pattern has been}%
\typeout{** loaded for the language `#1'. Using the pattern for}%
\typeout{** the default language instead.}%
\else
\language=\csname l@#1\endcsname
\fi
#2}}
\providecommand{\BIBdecl}{\relax}
\BIBdecl

\bibitem{Wolpert1997TEC}
D.~H. Wolpert and W.~G. Macready, ``No free lunch theorems for optimization,''
  \emph{IEEE Transactions on Evolutionary Computation}, vol.~1, no.~1, pp.
  67--82, 1997.

\bibitem{Lou2016MC}
Y.~Lou and S.~Y. Yuen, ``Non-revisiting genetic algorithm with adaptive
  mutation using constant memory,'' \emph{Memetic Computing}, vol.~8, no.~3,
  pp. 189--210, 2016, doi:10.1007/s12293-015-0178-6.

\bibitem{Akay2012IS}
B.~Akay and D.~Karaboga, ``A modified artificial bee colony algorithm for
  real-parameter optimization,'' \emph{Information Sciences}, vol. 192, pp.
  120--142, 2012.

\bibitem{Gandomi2012NCA}
A.~H. Gandomi and X.-S. Yang, ``Evolutionary boundary constraint handling
  scheme,'' \emph{Neural Computing and Applications}, vol.~21, no.~6, pp.
  1449--1462, 2012.

\bibitem{Chu2011IS}
W.~Chu, X.~Gao, and S.~Sorooshian, ``Handling boundary constraints for particle
  swarm optimization in high-dimensional search space,'' \emph{Information
  Sciences}, vol. 181, no.~20, pp. 4569--4581, 2011.

\bibitem{Zambrano2013CEC}
M.~Zambrano-Bigiarini, M.~Clerc, and R.~Rojas, ``Standard particle swarm
  optimisation 2011 at {CEC}-2013: {A} baseline for future {PSO}
  improvements,'' in \emph{IEEE Congress on Evolutionary Computation
  (CEC)}.\hskip 1em plus 0.5em minus 0.4em\relax IEEE, 2013, pp. 2337--2344.

\bibitem{Yuen2009TEC}
S.~Y. Yuen and C.~K. Chow, ``A genetic algorithm that adaptively mutates and
  never revisits,'' \emph{IEEE Transactions on Evolutionary Computation},
  vol.~13, no.~2, pp. 454--472, 2009.

\bibitem{Chow2010CEC}
C.~K. Chow and S.~Y. Yuen, ``Continuous non-revisiting genetic algorithm with
  random search space re-partitioning and one-gene-flip mutation,'' in
  \emph{IEEE Congress on Evolutionary Computation (CEC)}.\hskip 1em plus 0.5em
  minus 0.4em\relax IEEE, 2010, pp. 1--8.

\bibitem{Chow2012CEC}
------, ``Continuous non-revisiting genetic algorithm with overlapped search
  sub-region,'' in \emph{IEEE Congress on Evolutionary Computation
  (CEC)}.\hskip 1em plus 0.5em minus 0.4em\relax IEEE, 2012, pp. 1--8.

\bibitem{Hansen2006TNEC}
N.~Hansen, ``The {CMA} evolution strategy: {A} comparing review,'' in
  \emph{Towards a New Evolutionary Computation}.\hskip 1em plus 0.5em minus
  0.4em\relax Springer, 2006, pp. 75--102.

\bibitem{Hansen2016arXiv}
------, ``The {CMA} evolution strategy: {A} tutorial,'' \emph{arXiv preprint
  arXiv:1604.00772}, 2016.

\bibitem{Loshchilov2013CEC}
I.~Loshchilov, ``{CMA-ES} with restarts for solving {CEC} 2013 benchmark
  problems,'' in \emph{IEEE Congress on Evolutionary Computation (CEC)}.\hskip
  1em plus 0.5em minus 0.4em\relax IEEE, 2013, pp. 369--376.

\bibitem{Preuss2010GECCO}
M.~Preuss, ``Niching the {CMA-ES} via nearest-better clustering,'' in
  \emph{Genetic and Evolutionary Computation Conference}.\hskip 1em plus 0.5em
  minus 0.4em\relax ACM, 2010, pp. 1711--1718.

\bibitem{Shir2010EC}
O.~M. Shir, M.~Emmerich, and T.~B{\"a}ck, ``Adaptive niche radii and niche
  shapes approaches for niching with the {CMA-ES},'' \emph{Evolutionary
  Computation}, vol.~18, no.~1, pp. 97--126, 2010.

\bibitem{CEC2013}
J.~Liang, B.~Qu, P.~Suganthan, and A.~G. Hern{\'a}ndez-D{\'\i}az, ``Problem
  definitions and evaluation criteria for the {CEC} 2013 special session on
  real-parameter optimization,'' Zhengzhou University, China and Nanyang
  Technological University, Singapore, Tech. Rep., 2013.

\bibitem{CEC2017}
N.~Awad, M.~Ali, J.~Liang, B.~Qu, and P.~N. Suganthan, ``Problem definitions
  and evaluation criteria for the {CEC} 2017 special session and competition on
  single objective real-parameter numerical optimization,'' Nanyang
  Technological University, Singapore, Jordan University of Science and
  Technology, Jordan, Zhengzhou University, China, Tech. Rep., 2017.

\bibitem{Lou2018GECCO}
Y.~Lou, S.~Y. Yuen, and G.~Chen, ``Evolving benchmark functions using
  {K}ruskal-{W}allis test,'' in \emph{Genetic and Evolutionary Computation
  Conference (GECCO)}.\hskip 1em plus 0.5em minus 0.4em\relax ACM, 2018, pp.
  1337--1341.

\bibitem{CMA_Matlab}
\BIBentryALTinterwordspacing
{CMA-ES} source code. [Online]. Available:
  \url{http://cma.gforge.inria.fr/cmaes.m}
\BIBentrySTDinterwordspacing

\bibitem{NR_Matlab}
\BIBentryALTinterwordspacing
{cNrGA} source code. [Online]. Available:
  \url{http://www.ee.cityu.edu.hk/~syyuen/Public/Code.html}
\BIBentrySTDinterwordspacing

\bibitem{Yuen2016ASC}
S.~Y. Yuen, C.~K. Chow, X.~Zhang, and Y.~Lou, ``Which algorithm should {I}
  choose: {A}n evolutionary algorithm portfolio approach,'' \emph{Applied Soft
  Computing}, vol.~40, pp. 654--673, 2016.

\end{thebibliography}

%\section*{Appendix}
%\label{sec:apdx}

% --- Table #3+#4 CEC 2013 10D --- %
\begin{table*}[htbp]
	\begin{center}
		\caption{The detailed results of HR-CMA-ES and CMA-ES on solving CEC 2013 10D problems.}
		\label{tbres_cec13_10d}
		\begin{tabular}{c|ccccc|ccccc}
			\hline
			\multirow{2}{*}{\begin{tabular}[c]{@{}c@{}}CEC\\ 2013\end{tabular}} & \multicolumn{5}{c|}{HR-CMA-ES} & \multicolumn{5}{c}{CMA-ES}	\\ \cline{2-11}
			& best & worst & median & mean & std & best & worst & median & mean & std \\ \hline
			f1301 & 0.00E+00 & 2.27E-13 & 2.27E-13 & 1.74E-13 & 9.78E-14 & 0.00E+00 & 2.27E-13 & 2.27E-13 & 1.52E-13 & 1.09E-13 \\
			f1302 & 0.00E+00 & 2.27E-13 & 2.27E-13 & 1.89E-13 & 8.62E-14 & 0.00E+00 & 2.27E-13 & 2.27E-13 & 1.97E-13 & 7.86E-14 \\
			f1303 & 2.27E-13 & 6.82E-13 & 2.27E-13 & 3.41E-13 & 1.55E-13 & 2.27E-13 & 9.09E-13 & 2.27E-13 & 3.71E-13 & 1.84E-13 \\
			f1304 & 0.00E+00 & 2.27E-13 & 2.27E-13 & 2.20E-13 & 4.15E-14 & 0.00E+00 & 2.27E-13 & 2.27E-13 & 1.97E-13 & 7.86E-14 \\
			f1305 & 4.55E-13 & 2.73E-12 & 9.09E-13 & 1.15E-12 & 5.38E-13 & 3.41E-13 & 2.73E-12 & 1.02E-12 & 1.13E-12 & 5.73E-13 \\ \hline
			f1306 & 0.00E+00 & 2.27E-13 & 1.14E-13 & 1.44E-13 & 6.63E-14 & 0.00E+00 & 2.27E-13 & 1.14E-13 & 1.14E-13 & 2.99E-14 \\
			f1307 & 4.40E-02 & 1.09E+01 & 4.15E-01 & 9.17E-01 & 2.02E+00 & 5.22E-03 & 4.55E+00 & 2.65E-01 & 7.73E-01 & 1.18E+00 \\
			f1308 & 2.03E+01 & 2.07E+01 & 2.05E+01 & 2.05E+01 & 1.18E-01 & 2.06E+01 & 2.11E+01 & 2.09E+01 & 2.08E+01 & 1.42E-01 \\
			f1309 & 2.88E-01 & 3.33E+00 & 1.67E+00 & 1.65E+00 & 8.26E-01 & 5.56E-01 & 3.34E+00 & 1.98E+00 & 1.94E+00 & 7.93E-01 \\
			f1310 & 0.00E+00 & 5.68E-14 & 5.68E-14 & 5.31E-14 & 1.44E-14 & 0.00E+00 & 1.14E-13 & 5.68E-14 & 5.68E-14 & 1.49E-14 \\
			f1311 & 1.14E-13 & 4.97E+00 & 1.99E+00 & 2.03E+00 & 1.02E+00 & 1.99E+00 & 7.96E+00 & 3.98E+00 & 4.48E+00 & 1.58E+00 \\
			f1312 & 9.95E-01 & 8.95E+00 & 4.97E+00 & 4.88E+00 & 1.82E+00 & 2.98E+00 & 6.96E+00 & 4.97E+00 & 4.58E+00 & 1.21E+00 \\
			f1313 & 1.17E+00 & 1.86E+01 & 8.64E+00 & 8.70E+00 & 4.19E+00 & 9.95E-01 & 1.72E+01 & 8.21E+00 & 7.95E+00 & 4.18E+00 \\
			f1314 & 3.66E+00 & 2.72E+02 & 9.50E+01 & 1.06E+02 & 7.23E+01 & 1.68E+01 & 8.18E+02 & 5.11E+02 & 4.82E+02 & 1.68E+02 \\
			f1315 & 1.02E+02 & 6.75E+02 & 4.27E+02 & 4.27E+02 & 1.60E+02 & 1.69E+02 & 7.72E+02 & 4.82E+02 & 4.51E+02 & 1.58E+02 \\
			f1316 & 3.69E-02 & 1.89E+00 & 9.80E-01 & 1.01E+00 & 6.20E-01 & 6.28E-02 & 3.69E+00 & 2.37E+00 & 2.04E+00 & 1.16E+00 \\
			f1317 & 6.20E+00 & 1.63E+01 & 1.25E+01 & 1.18E+01 & 2.61E+00 & 4.93E+00 & 1.88E+01 & 1.42E+01 & 1.35E+01 & 3.27E+00 \\
			f1318 & 2.73E+00 & 1.70E+01 & 1.39E+01 & 1.18E+01 & 4.21E+00 & 6.52E+00 & 1.84E+01 & 1.47E+01 & 1.45E+01 & 2.79E+00 \\
			f1319 & 1.88E-01 & 7.13E-01 & 5.11E-01 & 4.87E-01 & 1.26E-01 & 3.42E-01 & 7.37E-01 & 5.40E-01 & 5.15E-01 & 9.13E-02 \\
			f1320 & 2.63E+00 & 4.02E+00 & 3.62E+00 & 3.57E+00 & 3.56E-01 & 2.12E+00 & 4.34E+00 & 3.79E+00 & 3.69E+00 & 4.20E-01 \\ \hline
			f1321 & 2.00E+02 & 4.00E+02 & 2.00E+02 & 2.07E+02 & 3.66E+01 & 1.00E+02 & 2.00E+02 & 2.00E+02 & 1.57E+02 & 5.04E+01 \\
			f1322 & 4.10E+01 & 5.96E+02 & 2.39E+02 & 2.41E+02 & 1.28E+02 & 4.14E+02 & 8.78E+02 & 5.63E+02 & 6.00E+02 & 1.18E+02 \\
			f1323 & 1.36E+02 & 9.78E+02 & 5.72E+02 & 5.51E+02 & 1.77E+02 & 2.10E+02 & 9.03E+02 & 6.04E+02 & 5.62E+02 & 1.66E+02 \\
			f1324 & 2.00E+02 & 2.12E+02 & 2.07E+02 & 2.06E+02 & 3.77E+00 & 1.05E+02 & 1.37E+02 & 1.08E+02 & 1.11E+02 & 7.07E+00 \\
			f1325 & 2.00E+02 & 2.12E+02 & 2.03E+02 & 2.05E+02 & 4.09E+00 & 1.15E+02 & 2.10E+02 & 2.01E+02 & 1.99E+02 & 1.72E+01 \\
			f1326 & 1.01E+02 & 2.00E+02 & 1.06E+02 & 1.40E+02 & 4.67E+01 & 1.03E+02 & 1.14E+02 & 1.07E+02 & 1.08E+02 & 2.65E+00 \\
			f1327 & 3.00E+02 & 5.17E+02 & 3.02E+02 & 3.25E+02 & 5.62E+01 & 3.00E+02 & 3.09E+02 & 3.00E+02 & 3.01E+02 & 1.69E+00 \\
			f1328 & 3.00E+02 & 3.00E+02 & 3.00E+02 & 3.00E+02 & 3.04E-09 & 1.00E+02 & 3.00E+02 & 3.00E+02 & 2.67E+02 & 7.58E+01 \\ \hline
		\end{tabular}
	\end{center}
\end{table*}
% --- Table #3+#4 CEC 2013 10D --- %

% --- Table #5+#6 CEC 2013 30D --- %
\begin{table*}[htbp]
	\begin{center}
		\caption{The detailed results of HR-CMA-ES and CMA-ES on solving CEC 2013 30D problems.}
		\label{tbres_cec13_30d}
		\begin{tabular}{c|ccccc|ccccc}
			\hline
			\multirow{2}{*}{\begin{tabular}[c]{@{}c@{}}CEC\\ 2013\end{tabular}} & \multicolumn{5}{c|}{HR-CMA-ES} & \multicolumn{5}{c}{CMA-ES}	\\ \cline{2-11}
			& best & worst & median & mean & std & best & worst & median & mean & std \\ \hline
			f1301 & 2.27E-13 & 4.55E-13 & 4.55E-13 & 3.79E-13 & 1.09E-13  & 2.27E-13 & 4.55E-13 & 2.27E-13 & 3.26E-13 & 1.15E-13 \\
			f1302 & 4.55E-13 & 6.82E-13 & 4.55E-13 & 5.08E-13 & 9.78E-14  & 4.55E-13 & 6.82E-13 & 4.55E-13 & 5.00E-13 & 9.25E-14 \\
			f1303 & 6.83E-10 & 2.44E+02 & 3.85E-05 & 8.31E+00 & 4.45E+01  & 1.43E-11 & 2.59E+00 & 3.14E-06 & 1.54E-01 & 4.93E-01 \\
			f1304 & 4.55E-13 & 6.82E-13 & 4.55E-13 & 5.46E-13 & 1.13E-13  & 2.27E-13 & 6.82E-13 & 4.55E-13 & 5.00E-13 & 1.10E-13 \\
			f1305 & 2.61E-12 & 3.11E-10 & 1.07E-11 & 3.14E-11 & 6.28E-11  & 2.73E-12 & 2.75E-10 & 1.34E-11 & 2.69E-11 & 5.12E-11 \\ \hline
			f1306 & 2.27E-13 & 3.41E-13 & 2.27E-13 & 2.46E-13 & 4.31E-14  & 2.27E-13 & 3.41E-13 & 2.27E-13 & 2.50E-13 & 4.63E-14 \\
			f1307 & 1.01E+00 & 2.42E+01 & 1.08E+01 & 1.10E+01 & 6.59E+00  & 3.51E+00 & 3.36E+01 & 1.23E+01 & 1.38E+01 & 7.11E+00 \\
			f1308 & 2.09E+01 & 2.11E+01 & 2.10E+01 & 2.10E+01 & 6.98E-02  & 2.11E+01 & 2.13E+01 & 2.13E+01 & 2.13E+01 & 6.23E-02 \\
			f1309 & 1.05E+01 & 1.70E+01 & 1.40E+01 & 1.38E+01 & 1.65E+00  & 9.72E+00 & 1.72E+01 & 1.39E+01 & 1.38E+01 & 1.78E+00 \\
			f1310 & 1.14E-13 & 7.40E-03 & 1.71E-13 & 2.47E-04 & 1.35E-03  & 5.68E-14 & 7.40E-03 & 1.71E-13 & 2.47E-04 & 1.35E-03 \\
			f1311 & 8.95E+00 & 2.19E+01 & 1.39E+01 & 1.48E+01 & 3.94E+00  & 1.89E+01 & 3.98E+01 & 3.08E+01 & 3.00E+01 & 4.74E+00 \\
			f1312 & 1.69E+01 & 4.12E+01 & 2.96E+01 & 3.00E+01 & 5.58E+00  & 1.89E+01 & 3.28E+01 & 2.84E+01 & 2.78E+01 & 3.59E+00 \\
			f1313 & 4.32E+01 & 8.67E+01 & 6.39E+01 & 6.47E+01 & 1.11E+01  & 1.48E+01 & 9.36E+01 & 6.06E+01 & 6.00E+01 & 1.52E+01 \\
			f1314 & 7.68E+02 & 1.66E+03 & 1.18E+03 & 1.22E+03 & 2.67E+02  & 2.21E+03 & 3.33E+03 & 2.72E+03 & 2.73E+03 & 2.71E+02 \\
			f1315 & 1.45E+03 & 3.06E+03 & 2.49E+03 & 2.41E+03 & 3.84E+02  & 1.55E+03 & 2.95E+03 & 2.44E+03 & 2.43E+03 & 3.38E+02 \\
			f1316 & 2.22E-01 & 3.75E+00 & 3.13E+00 & 3.01E+00 & 6.63E-01  & 2.80E+00 & 6.01E+00 & 4.82E+00 & 4.77E+00 & 8.83E-01 \\
			f1317 & 2.94E+01 & 5.11E+01 & 4.14E+01 & 4.09E+01 & 5.26E+00  & 2.42E+01 & 6.86E+01 & 5.52E+01 & 5.06E+01 & 1.19E+01 \\
			f1318 & 3.32E+01 & 7.84E+01 & 6.12E+01 & 5.81E+01 & 1.25E+01  & 3.06E+01 & 9.01E+01 & 6.19E+01 & 5.94E+01 & 1.37E+01 \\
			f1319 & 1.34E+00 & 2.98E+00 & 2.33E+00 & 2.23E+00 & 3.85E-01  & 1.43E+00 & 2.96E+00 & 2.32E+00 & 2.27E+00 & 3.33E-01 \\
			f1320 & 1.19E+01 & 1.37E+01 & 1.32E+01 & 1.31E+01 & 5.06E-01  & 1.32E+01 & 1.50E+01 & 1.44E+01 & 1.44E+01 & 4.37E-01 \\ \hline
			f1321 & 1.00E+02 & 2.00E+02 & 2.00E+02 & 1.90E+02 & 3.05E+01  & 1.00E+02 & 2.00E+02 & 2.00E+02 & 1.80E+02 & 4.07E+01 \\
			f1322 & 5.82E+02 & 2.83E+03 & 1.15E+03 & 1.19E+03 & 4.91E+02  & 1.87E+03 & 3.62E+03 & 2.98E+03 & 2.96E+03 & 4.17E+02 \\
			f1323 & 2.06E+03 & 3.38E+03 & 2.76E+03 & 2.82E+03 & 3.57E+02  & 1.78E+03 & 3.25E+03 & 2.82E+03 & 2.73E+03 & 4.05E+02 \\
			f1324 & 2.25E+02 & 2.46E+02 & 2.39E+02 & 2.39E+02 & 6.15E+00  & 1.33E+02 & 2.30E+02 & 2.11E+02 & 1.95E+02 & 3.43E+01 \\
			f1325 & 2.44E+02 & 2.60E+02 & 2.54E+02 & 2.54E+02 & 4.27E+00  & 2.43E+02 & 2.59E+02 & 2.52E+02 & 2.52E+02 & 4.26E+00 \\
			f1326 & 2.00E+02 & 3.37E+02 & 2.84E+02 & 2.67E+02 & 5.82E+01  & 1.32E+02 & 2.00E+02 & 2.00E+02 & 1.88E+02 & 2.25E+01 \\
			f1327 & 5.60E+02 & 7.28E+02 & 6.80E+02 & 6.70E+02 & 4.20E+01  & 4.00E+02 & 6.57E+02 & 5.72E+02 & 5.64E+02 & 6.44E+01 \\
			f1328 & 1.00E+02 & 3.00E+02 & 3.00E+02 & 2.87E+02 & 5.07E+01  & 1.00E+02 & 3.00E+02 & 1.00E+02 & 1.41E+02 & 8.10E+01 \\ \hline
			
		\end{tabular}
	\end{center}
\end{table*}
% --- Table #5+#6 CEC 2013 30D --- %

% --- Table #7+#8 CEC 2017 10D --- %
\begin{table*}[htbp]
	\begin{center}
		\caption{The detailed results of HR-CMA-ES and CMA-ES on solving CEC 2017 10D problems.}
		\label{tbres_cec17_10d}
		\begin{tabular}{c|ccccc|ccccc}
			\hline
			\multirow{2}{*}{\begin{tabular}[c]{@{}c@{}}CEC\\ 2017\end{tabular}} & \multicolumn{5}{c|}{HR-CMA-ES} & \multicolumn{5}{c}{CMA-ES}	\\ \cline{2-11}
			& best & worst & median & mean & std & best & worst & median & mean & std \\ \hline
			f1701 & 0.00E+00 & 1.42E-14 & 1.42E-14 & 1.23E-14 & 4.91E-15 & 0.00E+00 & 1.42E-14 & 1.42E-14 & 1.18E-14 & 5.39E-15 \\
			f1703 & 0.00E+00 & 5.68E-14 & 5.68E-14 & 4.36E-14 & 2.45E-14 & 0.00E+00 & 5.68E-14 & 5.68E-14 & 4.55E-14 & 2.31E-14 \\ \hline
			f1704 & 0.00E+00 & 5.68E-14 & 5.68E-14 & 5.49E-14 & 1.04E-14 & 0.00E+00 & 5.68E-14 & 5.68E-14 & 5.31E-14 & 1.44E-14 \\
			f1705 & 2.98E+00 & 8.95E+00 & 5.97E+00 & 5.87E+00 & 1.68E+00 & 1.99E+00 & 8.95E+00 & 4.97E+00 & 5.50E+00 & 1.58E+00 \\
			f1706 & 1.39E-05 & 1.38E-03 & 9.72E-05 & 2.11E-04 & 3.02E-04 & 2.68E-05 & 3.17E-03 & 2.15E-04 & 4.50E-04 & 6.88E-04 \\
			f1707 & 4.63E+00 & 1.70E+01 & 1.29E+01 & 1.20E+01 & 3.26E+00 & 4.61E+00 & 1.60E+01 & 1.33E+01 & 1.20E+01 & 3.60E+00 \\
			f1708 & 2.98E+00 & 6.96E+00 & 4.97E+00 & 4.88E+00 & 1.24E+00 & 2.98E+00 & 8.95E+00 & 4.97E+00 & 5.14E+00 & 1.48E+00 \\
			f1709 & 0.00E+00 & 1.14E-13 & 1.14E-13 & 1.06E-13 & 2.88E-14 & 0.00E+00 & 1.14E-13 & 1.14E-13 & 1.10E-13 & 2.08E-14 \\
			f1710 & 1.02E+01 & 5.25E+02 & 2.66E+02 & 2.74E+02 & 1.56E+02 & 5.22E+01 & 7.45E+02 & 4.35E+02 & 3.81E+02 & 1.59E+02 \\ \hline
			f1711 & 1.99E+00 & 2.19E+01 & 6.96E+00 & 9.06E+00 & 5.76E+00 & 3.98E+00 & 4.38E+01 & 1.19E+01 & 1.41E+01 & 8.43E+00 \\
			f1712 & 6.82E-13 & 4.40E+02 & 1.58E+02 & 1.79E+02 & 1.16E+02 & 1.14E-12 & 5.06E+02 & 1.61E+02 & 1.87E+02 & 1.52E+02 \\
			f1713 & 1.31E+01 & 2.22E+02 & 7.67E+01 & 8.27E+01 & 4.15E+01 & 2.80E+01 & 4.20E+02 & 1.06E+02 & 1.28E+02 & 8.90E+01 \\
			f1714 & 2.10E+01 & 5.18E+01 & 2.99E+01 & 3.15E+01 & 7.54E+00 & 2.20E+01 & 6.48E+01 & 3.64E+01 & 3.73E+01 & 1.11E+01 \\
			f1715 & 2.31E+00 & 8.47E+01 & 2.39E+01 & 2.75E+01 & 1.96E+01 & 2.00E+00 & 7.57E+01 & 2.75E+01 & 3.34E+01 & 2.14E+01 \\
			f1716 & 9.33E-01 & 1.22E+02 & 1.99E+00 & 1.05E+01 & 2.34E+01 & 1.08E+00 & 2.20E+02 & 2.19E+00 & 3.62E+01 & 6.50E+01 \\
			f1717 & 1.69E+01 & 5.52E+01 & 2.88E+01 & 3.05E+01 & 8.55E+00 & 2.10E+01 & 6.12E+01 & 3.05E+01 & 3.36E+01 & 1.03E+01 \\
			f1718 & 2.00E+01 & 1.02E+02 & 2.67E+01 & 3.36E+01 & 1.78E+01 & 2.03E+01 & 9.96E+01 & 3.28E+01 & 4.45E+01 & 2.46E+01 \\
			f1719 & 4.80E+00 & 3.19E+01 & 9.97E+00 & 1.31E+01 & 7.82E+00 & 4.19E+00 & 3.27E+01 & 1.14E+01 & 1.25E+01 & 6.51E+00 \\
			f1720 & 2.10E+01 & 6.01E+01 & 3.16E+01 & 3.38E+01 & 1.03E+01 & 2.20E+01 & 1.65E+02 & 4.94E+01 & 5.62E+01 & 3.04E+01 \\ \hline
			f1721 & 1.91E+02 & 2.11E+02 & 2.07E+02 & 2.06E+02 & 4.43E+00 & 1.00E+02 & 2.10E+02 & 2.07E+02 & 2.03E+02 & 1.95E+01 \\
			f1722 & 1.00E+02 & 1.00E+02 & 1.00E+02 & 1.00E+02 & 2.89E-11 & 1.42E-09 & 1.00E+02 & 1.00E+02 & 7.49E+01 & 3.76E+01 \\
			f1723 & 3.05E+02 & 3.14E+02 & 3.09E+02 & 3.09E+02 & 2.02E+00 & 3.05E+02 & 3.10E+02 & 3.07E+02 & 3.07E+02 & 1.31E+00 \\
			f1724 & 3.23E+02 & 3.40E+02 & 3.35E+02 & 3.35E+02 & 3.39E+00 & 1.00E+02 & 3.37E+02 & 3.31E+02 & 2.82E+02 & 7.61E+01 \\
			f1725 & 3.98E+02 & 3.98E+02 & 3.98E+02 & 3.98E+02 & 1.71E-01 & 1.00E+02 & 3.98E+02 & 3.98E+02 & 3.78E+02 & 7.56E+01 \\
			f1726 & 2.00E+02 & 1.19E+03 & 3.00E+02 & 3.26E+02 & 1.65E+02 & 2.00E+02 & 3.00E+02 & 3.00E+02 & 2.77E+02 & 4.30E+01 \\
			f1727 & 3.89E+02 & 3.96E+02 & 3.91E+02 & 3.92E+02 & 2.17E+00 & 3.90E+02 & 3.98E+02 & 3.95E+02 & 3.95E+02 & 1.59E+00 \\
			f1728 & 3.00E+02 & 5.84E+02 & 3.00E+02 & 4.27E+02 & 1.41E+02 & 3.00E+02 & 3.28E+02 & 3.00E+02 & 3.01E+02 & 5.05E+00 \\
			f1729 & 2.42E+02 & 3.08E+02 & 2.58E+02 & 2.64E+02 & 1.97E+01 & 2.37E+02 & 2.98E+02 & 2.67E+02 & 2.68E+02 & 1.68E+01 \\
			f1730 & 3.95E+02 & 8.18E+05 & 4.70E+02 & 5.49E+04 & 2.07E+05 & 3.95E+02 & 5.55E+02 & 4.50E+02 & 4.57E+02 & 5.16E+01 \\ \hline
		\end{tabular}
	\end{center}
\end{table*}
% --- Table #7+#8 CEC 2017 10D --- %

% --- Table #9+#X CEC 2017 30D --- %
\begin{table*}[htbp]
	\begin{center}
		\caption{The detailed results of HR-CMA-ES and CMA-ES on solving CEC 2017 30D problems.}
		\label{tbres_cec17_30d}
		\begin{tabular}{c|ccccc|ccccc}
			\hline
			\multirow{2}{*}{\begin{tabular}[c]{@{}c@{}}CEC\\ 2017\end{tabular}} & \multicolumn{5}{c|}{HR-CMA-ES} & \multicolumn{5}{c}{CMA-ES}	\\ \cline{2-11}
			& best & worst & median & mean & std & best & worst & median & mean & std \\ \hline
			f1701 & 1.42E-14 & 2.84E-14 & 2.84E-14 & 2.56E-14 & 5.78E-15 & 1.42E-14 & 2.84E-14 & 2.84E-14 & 2.61E-14 & 5.39E-15 \\
			f1703 & 5.68E-14 & 1.71E-13 & 1.14E-13 & 1.08E-13 & 2.73E-14 & 5.68E-14 & 1.71E-13 & 1.14E-13 & 1.17E-13 & 2.56E-14 \\ \hline
			f1704 & 1.14E-13 & 5.86E+01 & 2.27E-13 & 8.36E+00 & 2.01E+01 & 1.14E-13 & 5.86E+01 & 1.71E-13 & 4.31E+00 & 1.48E+01 \\
			f1705 & 2.65E+01 & 4.38E+01 & 3.20E+01 & 3.33E+01 & 4.01E+00 & 2.29E+01 & 3.88E+01 & 3.08E+01 & 3.09E+01 & 4.71E+00 \\
			f1706 & 9.26E-04 & 1.38E-01 & 1.55E-02 & 2.73E-02 & 3.30E-02 & 4.78E-04 & 5.49E-01 & 9.27E-03 & 4.26E-02 & 1.03E-01 \\
			f1707 & 3.00E+01 & 6.33E+01 & 5.25E+01 & 5.20E+01 & 8.59E+00 & 3.30E+01 & 6.51E+01 & 5.53E+01 & 5.19E+01 & 9.19E+00 \\
			f1708 & 2.19E+01 & 4.28E+01 & 3.26E+01 & 3.25E+01 & 5.64E+00 & 2.29E+01 & 4.08E+01 & 3.33E+01 & 3.30E+01 & 4.63E+00 \\
			f1709 & 1.14E-13 & 8.95E-02 & 3.41E-13 & 8.95E-03 & 2.73E-02 & 2.27E-13 & 8.95E-02 & 3.41E-13 & 5.97E-03 & 2.27E-02 \\
			f1710 & 1.65E+03 & 3.06E+03 & 2.48E+03 & 2.38E+03 & 3.75E+02 & 1.87E+03 & 3.12E+03 & 2.58E+03 & 2.53E+03 & 3.64E+02 \\ \hline
			f1711 & 4.08E+01 & 1.18E+02 & 7.31E+01 & 7.75E+01 & 1.78E+01 & 4.08E+01 & 1.47E+02 & 9.45E+01 & 9.64E+01 & 2.45E+01 \\
			f1712 & 1.64E+02 & 1.64E+03 & 1.02E+03 & 1.05E+03 & 4.12E+02 & 1.69E+02 & 1.71E+03 & 9.67E+02 & 9.90E+02 & 3.15E+02 \\
			f1713 & 7.13E+02 & 2.66E+03 & 1.51E+03 & 1.57E+03 & 4.98E+02 & 7.02E+02 & 2.42E+03 & 1.53E+03 & 1.54E+03 & 5.34E+02 \\
			f1714 & 7.06E+01 & 2.19E+02 & 1.33E+02 & 1.42E+02 & 3.92E+01 & 8.17E+01 & 2.25E+02 & 1.57E+02 & 1.54E+02 & 4.10E+01 \\
			f1715 & 1.12E+02 & 6.46E+02 & 2.90E+02 & 3.12E+02 & 1.43E+02 & 8.19E+01 & 5.78E+02 & 2.85E+02 & 2.91E+02 & 1.18E+02 \\
			f1716 & 1.24E+01 & 1.02E+03 & 3.81E+02 & 4.13E+02 & 2.24E+02 & 5.33E+01 & 8.53E+02 & 4.77E+02 & 4.77E+02 & 2.07E+02 \\
			f1717 & 5.54E+01 & 3.53E+02 & 1.13E+02 & 1.47E+02 & 8.09E+01 & 5.85E+01 & 4.04E+02 & 2.15E+02 & 2.16E+02 & 9.41E+01 \\
			f1718 & 6.00E+01 & 2.85E+02 & 1.59E+02 & 1.64E+02 & 6.20E+01 & 8.40E+01 & 3.30E+02 & 1.75E+02 & 1.72E+02 & 5.94E+01 \\
			f1719 & 5.92E+01 & 2.42E+02 & 1.38E+02 & 1.44E+02 & 4.56E+01 & 5.67E+01 & 3.01E+02 & 1.51E+02 & 1.55E+02 & 4.93E+01 \\
			f1720 & 1.12E+02 & 4.88E+02 & 2.42E+02 & 2.56E+02 & 1.05E+02 & 1.15E+02 & 5.14E+02 & 2.31E+02 & 2.67E+02 & 1.21E+02 \\ \hline
			f1721 & 2.19E+02 & 2.46E+02 & 2.35E+02 & 2.35E+02 & 6.95E+00 & 2.25E+02 & 2.44E+02 & 2.34E+02 & 2.34E+02 & 4.93E+00 \\
			f1722 & 1.00E+02 & 3.20E+03 & 1.00E+02 & 3.88E+02 & 8.84E+02 & 1.00E+02 & 3.20E+03 & 2.47E+03 & 1.93E+03 & 1.16E+03 \\
			f1723 & 3.75E+02 & 4.13E+02 & 3.93E+02 & 3.92E+02 & 9.03E+00 & 3.67E+02 & 3.99E+02 & 3.83E+02 & 3.83E+02 & 8.60E+00 \\
			f1724 & 4.46E+02 & 4.77E+02 & 4.61E+02 & 4.61E+02 & 7.23E+00 & 2.00E+02 & 4.58E+02 & 4.49E+02 & 4.40E+02 & 4.56E+01 \\
			f1725 & 3.83E+02 & 3.87E+02 & 3.83E+02 & 3.84E+02 & 1.55E+00 & 3.83E+02 & 3.87E+02 & 3.83E+02 & 3.84E+02 & 1.55E+00 \\
			f1726 & 1.10E+03 & 1.54E+03 & 1.44E+03 & 1.42E+03 & 9.25E+01 & 2.00E+02 & 1.41E+03 & 1.26E+03 & 1.09E+03 & 4.14E+02 \\
			f1727 & 5.01E+02 & 5.44E+02 & 5.15E+02 & 5.17E+02 & 1.04E+01 & 5.02E+02 & 5.36E+02 & 5.21E+02 & 5.19E+02 & 8.44E+00 \\
			f1728 & 3.00E+02 & 3.00E+02 & 3.00E+02 & 3.00E+02 & 1.49E-10 & 3.00E+02 & 3.00E+02 & 3.00E+02 & 3.00E+02 & 9.32E-11 \\
			f1729 & 5.16E+02 & 9.78E+02 & 7.35E+02 & 7.36E+02 & 1.19E+02 & 5.12E+02 & 9.67E+02 & 7.31E+02 & 7.38E+02 & 1.19E+02 \\
			f1730 & 2.12E+03 & 2.74E+03 & 2.41E+03 & 2.41E+03 & 1.76E+02 & 2.13E+03 & 2.87E+03 & 2.46E+03 & 2.49E+03 & 2.04E+02 \\ \hline
		\end{tabular}
	\end{center}
\end{table*}
% --- Table #9+#X CEC 2017 30D --- % 

\end{document}